\documentclass[conference]{IEEEtran}
\IEEEoverridecommandlockouts

\usepackage{booktabs}
\usepackage{times}
\usepackage{soul}
\usepackage{url}
\usepackage[hidelinks]{hyperref}
\usepackage[utf8]{inputenc}
\usepackage[small]{caption}

\usepackage{amsmath}
\usepackage{amsthm}
\usepackage{booktabs}
\usepackage{multirow}

\usepackage{graphicx}

\usepackage{subcaption}
\usepackage{mwe}
\usepackage{algorithm}
\usepackage{algorithmic}
\usepackage{bm}

\usepackage{enumitem}
\usepackage{amsfonts}
\usepackage{algorithmic}
\usepackage{textcomp}
\usepackage{xcolor}
\def\BibTeX{{\rm B\kern-.05em{\sc i\kern-.025em b}\kern-.08em
    T\kern-.1667em\lower.7ex\hbox{E}\kern-.125emX}}
\begin{document}

\title{Heterogeneous Similarity Graph Neural Network on Electronic Health Records}

\author{
    \IEEEauthorblockN{
    Zheng Liu\IEEEauthorrefmark{1}, 
    Xiaohan Li\IEEEauthorrefmark{1},
    Hao Peng\IEEEauthorrefmark{2}, 
    Lifang He\IEEEauthorrefmark{3},  
    Philip S. Yu\IEEEauthorrefmark{1}}
    \IEEEauthorblockA{\IEEEauthorrefmark{1}University of Illinois at Chicago, Chicago, IL, USA
    \\\{zliu212, xli241, psyu\}@uic.edu}
    \IEEEauthorblockA{\IEEEauthorrefmark{2}Beihang University, Beijing, China
    \\penghao@act.buaa.edu.cn}
    \IEEEauthorblockA{\IEEEauthorrefmark{3}Lehigh University, Bethlehem, PA, USA
    \\lih319@lehigh.edu}
}

\IEEEoverridecommandlockouts
\IEEEpubid{\makebox[\columnwidth]{978-1-7281-6251-5/20/\$31.00 ~\copyright2020 IEEE \hfill} \hspace{\columnsep}\makebox[\columnwidth]{ }}

\maketitle
\IEEEpubidadjcol

\begin{abstract}
Mining Electronic Health Records (EHRs) becomes a promising topic because of the rich information they contain. By learning from EHRs, machine learning models can be built to help human expert to make medical decisions and thus improve healthcare quality. Recently, many models based on sequential or graph model are proposed to achieve this goal. EHRs contain multiple entities and relations, and can be viewed as a heterogeneous graph. However, previous studies ignore the heterogeneity in EHRs. On the other hand, current heterogeneous graph neural networks cannot be simply used on EHR graph because of the existence of hub nodes in it. To address this issue, we propose  Heterogeneous Similarity Graph Neural Network (\textbf{\textit{HSGNN}}) to analyze EHRs with a novel heterogeneous GNN. Our framework consists of two parts: one is a preprocessing method and the other is an end-to-end GNN. The preprocessing method normalizes edges and splits the EHR graph into multiple homogeneous graphs while each homogeneous graph contains partial information of the original EHR graph. The GNN takes all homogeneous graphs as input and fuses all of them into one graph to make prediction. Experimental results show that HSGNN outperforms other baselines in the diagnosis prediction task.

\end{abstract}

\section{Introduction}
The accumulation of large-scale Electronic Health Records (EHRs) provides us with great opportunity of deep learning applications on healthcare. Recently, many deep learning models have been applied to medical tasks such as phenotyping \cite{DBLP:conf/ijcai/FuHXS19,bai2018ehr}, medical predictive modeling \cite{DBLP:conf/kdd/ZhangTDZW19,lipton2015learning} and medication recommendation \cite{DBLP:conf/aaai/ShangXMLS19}. 

\begin{figure}[t]
\centering
\includegraphics[width=0.8\columnwidth]{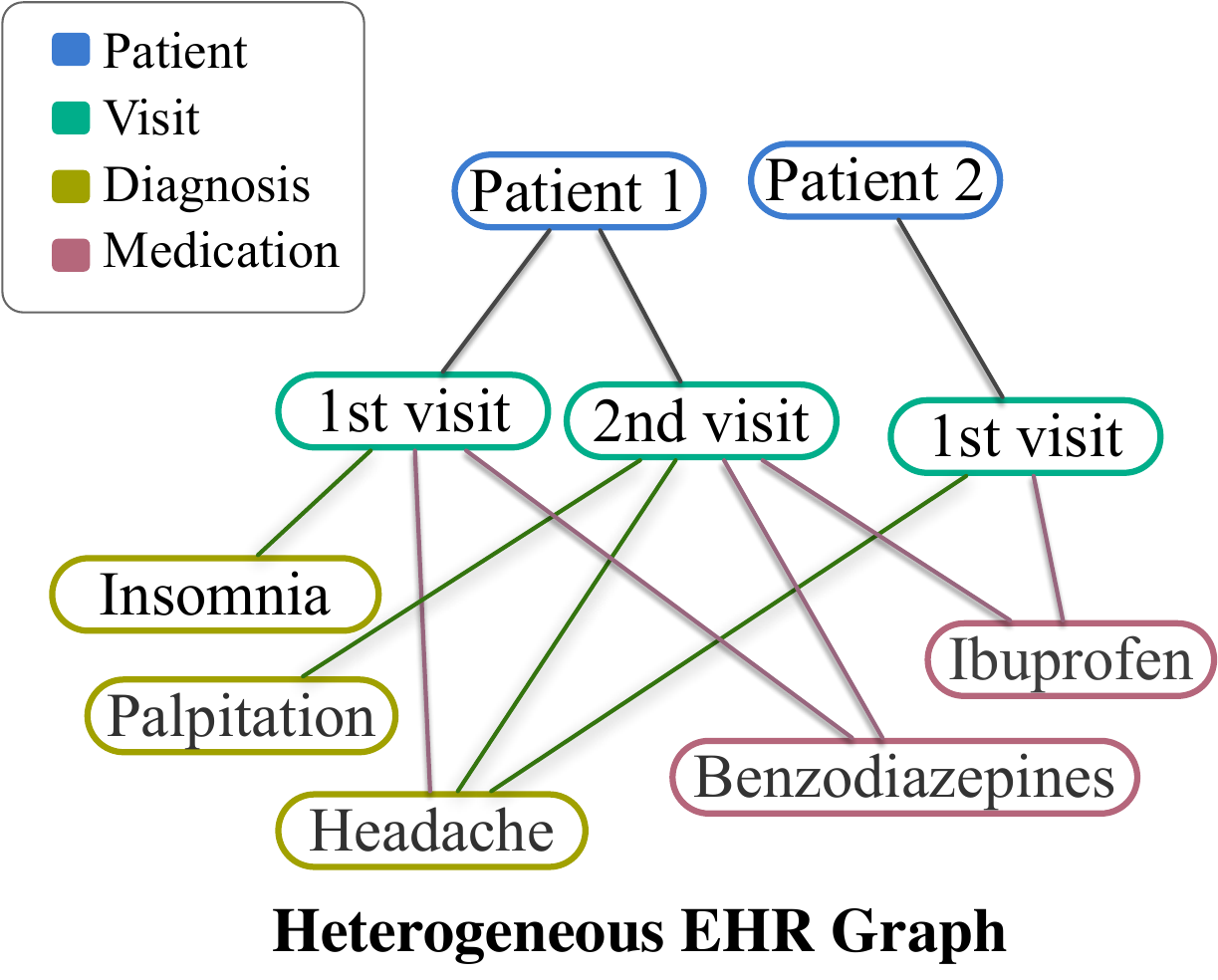} 
\caption{An example of heterogeneous EHR graph.}
\label{hin}
\end{figure}

Generally, raw EHRs consist of multiple kinds of features of patients, including demographics, observations, diagnoses, medications, and procedures ordered by time. For example, Fig.\ref{hin} shows an example of an EHR graph with two patients and three visit records. In Fig.\ref{hin}, there are two patients $p_1$ and $p_2$, where $p_1$ has visited the medical provider twice and $p_2$ has visited once (with timestamp recorded). During the visit some diagnoses or medications may occur to the patient. All medical concepts such as diagnosis, medications and procedures are medical codes and scientists can easily track them through some medical ontology. Because one patient can have multiple visits recorded, EHR can be viewed as sequential historical records for each patient. 
Moreover, because of the variety of medical codes and their relations, EHR can be viewed as a heterogeneous graph with multiple types of nodes and edges.

EHR analysis plays an important role in medical research and can improve the level of healthcare. By learning from EHRs, scientists can either discover useful facts or build intelligent applications. For example, the prescriptions in EHRs can help make medication recommendations \cite{DBLP:conf/aaai/ShangXMLS19}, and the phenotypes of patients indicate the distribution of cohorts \cite{DBLP:conf/icdm/CheL17}. With Artificial Intelligence (AI) technologies, scientists can build applications to provide useful suggestions to doctors, or let patients understand their physical conditions better. 


To build such a medical AI application, a key issue is to learn effective representations for each medical concept and patient \cite{cai2018medical, choi2016multi}. However, there are two challenges of learning such representations. One is data insufficiency. Due to the the privacy policy and the expense of collecting data, the volume of an EHR dataset is generally smaller than image or language datasets. Therefore, it is difficult for deep learning models designed for images or languages tasks to process EHR data. 
The other is the heterogeneity of EHR. 
EHR is of complex structure and contains multiple relationships. Only when all relations are properly used then the model can achieve a satisfactory performance.

Previously, many models regard EHRs as sequences and use sequential models such as RNNs to analyze EHR
\cite{DBLP:conf/nips/ChoiBSKSS16,DBLP:conf/kdd/MaCZYSG17,DBLP:journals/corr/LiptonKEW15,DBLP:conf/mlhc/ChoiBSSS16,DBLP:journals/corr/AczonLHGFWW17}. These methods use historical information to predict the next-period situation of a patient. However, sequential models are not enough to capture structural information and need a large amount of data to train.

To address these issues, some other approaches take EHR as a graph shown in Fig.\ref{hin}, and then use graph neural networks (GNNs) to learn embedding vectors for each node \cite{DBLP:conf/cikm/MaYXCZG18,choilearning,DBLP:conf/kdd/ChoiBSSS17,DBLP:conf/nips/ChoiXSS18}.
Among them, GRAM \cite{DBLP:conf/kdd/ChoiBSSS17} proposes the first graph-based model that can integrate external hierarchical ontologies when generating results. MiME \cite{DBLP:conf/nips/ChoiXSS18} learns multi-level representations of medical codes based on EHR data in a hierarchical order. Graph convolutional transformer (GCT) \cite{choilearning} 
learns the medical representations together with the hidden causal structure of EHR using the ``pre-training\&fine-tuning" procedure. Compared with sequential models, these graph-based models are more robust to insufficient data because  of the use of structural information: the model can use neighbor information to complete missing entries in the dataset.

As the above GNN models are designed only for homogeneous graphs, they fail to take all kinds of medical codes into account. EHR data contains multiple kinds of medical codes and relations, so it is naturally heterogeneous. To capture multiple relations in the graph, heterogeneous graph neural networks \cite{DBLP:conf/www/WangJSWYCY19,DBLP:conf/kdd/ZhangSHSC19,DBLP:conf/www/0004ZMK20} are necessary in the EHR analysis. Basically, these models take a heterogeneous graph as input and process different kinds of nodes or meta-paths\cite{DBLP:journals/pvldb/SunHYYW11} respectively. 

However, applying these models on EHR graphs directly can cause very low performance because of the hub nodes with high visibility \cite{DBLP:conf/kdd/ShiCZG017}. For example, if an EHR graph contains the ``gender" information, and then all patient nodes would link to either ``male" or ``female" nodes. If we don't conduct a normalization on these links, these two gender nodes would strongly influence all other nodes. After applying heterogeneous GNNs, all other nodes in the graph will eventually learn the same representations as either ``male" or ``female" nodes. This phenomenon is similar to the
over-smoothing \cite{DBLP:conf/aaai/LiHW18} problem. Over-smoothing means after applying GNNs with multiple layers, all node embeddings become close and finally indistinguishable. Some studies \cite{DBLP:conf/aaai/ChenLLLZS20} indicate the reason of over-smoothing is the existence of noise in the graph, which can be supported in our case: since gender is not the most informative attribute of a patient (containing too much noise), introducing it into the graph does not always helpful to the prediction task.

To address this issue, we propose \textbf{H}eterogeneous \textbf{S}imilarity \textbf{G}raph \textbf{N}eural \textbf{N}etwork (HSGNN), a framework using GNN to analyze EHR graphs. It consists of two parts: the preprocessing step and the end-to-end model. In the preprocessing step, we first construct the heterogeneous EHR graph, and then split it into multiple homogeneous subgraphs according to the  weight assigned to each edge. By doing so, we eliminate the noise in the original heterogeneous graph while preserving its structural information. After preprocessing step,  each subgraph contains partial information of the original graph. Then in the end-to-end model, we try to combine all subgraphs together into one integrated homogeneous graph $A_{meta}$ so that it can be input into any general GNN layers to make downstream predictions. Inspired by \cite{choilearning}, we set all weights in $A_{meta}$ as trainable variables but not fixed values. It means all weights in $A_{meta}$ are randomly initialized before training, and are optimized during the model training process.

Compared with previous models, HSGNN has these innovations: First, to the best of our knowledge, this is the first study that uses heterogeneous graph structure to represent EHR data, which can preserve the most information. Second, in the preprocessing step, HSGNN uses similarity values to represent the weights in the graph. This method is proved effective in the experiments to reduce over-smoothing. Third, we use trainable weights and construct a new graph in HSGNN, which can reveal true relationship between each nodes.

To demonstrate the advantages of HSGNN, we evaluate its performance on the MIMIC-III dataset. On the diagnosis prediction task, HSGNN outperforms all other baseline approaches and achieves state-of-the-art performance. We also prove the effectiveness of using similarity values by comparing HSGNN with a variant that uses $PathCount$s as graph weights. Finally, we visualize the structure of learned graph to prove that HSGNN can learn a new graph with higher quality. Conclusively, we make the following contributions in this paper:
\begin{itemize}
    \item We propose a novel framework HSGNN, which can learn informative representations for medical codes and make predictions for patients in EHR.
    \item We use the similarity subgraphs generated from original heterogeneous graph as input, which is shown effective to improve the performance of prediction.
    \item We propose an end-to-end model that can jointly learn high-quality graph embeddings based on similarity subgraphs and make accurate predictions .
    \item Experimental results show the superiority of our proposed model on the diagnosis prediction task. Experiments also prove the effectiveness of using similarity subgraphs and the quality of learned graph embeddings.
\end{itemize}

The code of our proposed HSGNN is available at \url{https://github.com/ErikaLiu/HSGNN}.

\section{Related Works}
Since EHR analysis is an interdisciplinary topic, many studies are related to our work. In this section, we only choose the most representative and inspiring studies. These studies mainly focus on four aspects: 1. Graph Neural Networks, 2. GNN-based EHR analysis, 3. heterogeneous graph neural networks and 4. some studies of the nature of graph.

\subsection{Graph Neural Networks}
Currently, Graph Neural Networks (GNNs) have been widely explored to process graph-structure data. Motivated by convolutional neural networks, 
Bruna et al. \cite{bruna2013spectral} propose graph convolutions in spectral domain. Then, Kipf and Welling \cite{DBLP:conf/iclr/KipfW17} simplified the previous graph convolution operation and designed a Graph Convolutional Network (GCN) model. Besides, to inductively generate node embeddings, Hamilton et al. propose the GraphSAGE~\cite{hamilton2017inductive} model to learn node embeddings with sampling and aggregation functions. All these models have shown their performance on many tasks  \cite{dou2020enhancing, peng2019fine, liu2020basket, li2020dynamic, gao2020hincti}.

\subsection{GNN-based EHR analysis}
Previously, many studies use RNNs to analyse EHR \cite{DBLP:conf/nips/ChoiBSKSS16,DBLP:conf/kdd/MaCZYSG17,DBLP:journals/corr/LiptonKEW15}. However, with the improvement of graph neural networks \cite{DBLP:conf/iclr/KipfW17,DBLP:journals/corr/abs-1710-10903,hamilton2017inductive}, many approaches develop GNNs to analyse EHR \cite{DBLP:conf/cikm/MaYXCZG18,choilearning,DBLP:conf/kdd/ChoiBSSS17,cao2020multi}. These models can capture structural information from raw EHR and thus outperform previous approaches.

Among these models, GRAM \cite{DBLP:conf/kdd/ChoiBSSS17} and KAME \cite{DBLP:conf/cikm/MaYXCZG18} use GNNs to process external hierarchical ontologies. They can learn embeddings for medical codes in the ontologies and then these embeddings can be used for downstream tasks. MiME \cite{DBLP:conf/nips/ChoiXSS18} and GCT \cite{choilearning} assume that there are some  latent causal relations between different kinds of medical codes in EHR. Based on this assumption, MiME learns multilevel representations in a hierarchical order and GCT can jointly learn the hidden causal structure of EHR while performing predictions. Above studies only focus on homogeneous graphs, while raw EHRs contain multiple kinds of medical codes and thus are naturally heterogeneous. This fact provides us with opportunities to model EHR with heterogeneous graphs.

\subsection{Heterogeneous Graph Neural Networks}
According to \cite{DBLP:journals/tkde/ShiLZSY17}, a heterogeneous information network (HIN) is an information network with multiple kinds of nodes and edges. To process HIN, a key issue is to deal with the heterogeneity of the network. Here we introduce some methods in the previous studies to eliminate the heterogeneity of the network.

HAN \cite{DBLP:conf/www/WangJSWYCY19} is the first study using graph attention network to process heterogeneous graphs. MAGNN \cite{DBLP:conf/www/0004ZMK20} is another recent study proposing aggregators to make inductive learning on heterogeneous graphs. Both of these two models use meta-path when processing heterogeneous graphs since it can capture meaningful patterns in the graph. Also, both of their models consists of two modules: the meta-path level GNN and the node level GNN, which can aggregate node features hierarchically. HetGNN \cite{DBLP:conf/kdd/ZhangSHSC19} proposes another method to eliminate the heterogeneity, which uses random walk and type-based aggregators. However, in the experiment we 
find that these methods do not perform ideally because they did not deal with nodes with different visibility properly.

\subsection{Over-smoothing and Node Visibility}
According to \cite{DBLP:conf/aaai/LiHW18}, after applying GNN with multiple layers, the derived node embeddings will become closer to each other and finally indistinguishable. This is called over-smoothing and \cite{DBLP:conf/aaai/LiHW18} is the first work discover this phenomenon.

Recently, the causes of this phenomenon are still being investigated and some studies are trying to resolve it. For example, \cite{DBLP:conf/iclr/ZhaoA20} discovers row-level and col-level over-smoothing because information wrongly spread through nodes and features. Another study \cite{DBLP:conf/aaai/ChenLLLZS20} attribute over-smoothing to the noise in the network. Nevertheless, these different explanations may direct to the same reason. That is, the structural information of the graph may not accurate enough, making information spread to wrong nodes or wrong features through GNN. Therefore, correct the ``wrong edges" in the graph is a possible way to overcome over-smoothing.

On the other hand, many traditional studies focus more on the nature of graph \cite{DBLP:journals/pvldb/SunHYYW11,DBLP:conf/kdd/ShiCZG017}. These research propose the concept  ``node visibility" to measure the influence of one node on the whole graph. Generally, the degree of the node can be used to measure the visibility of it. If one node have many neighbors, it can influence more other nodes and making itself ``visible" in the whole graph. In GNNs, the existence of these highly visible nodes is one reason of over-smoothing because they can result in  
multiple nodes having similar embeddings.

\begin{figure*}[t]
\centering
\includegraphics[width=2.05\columnwidth]{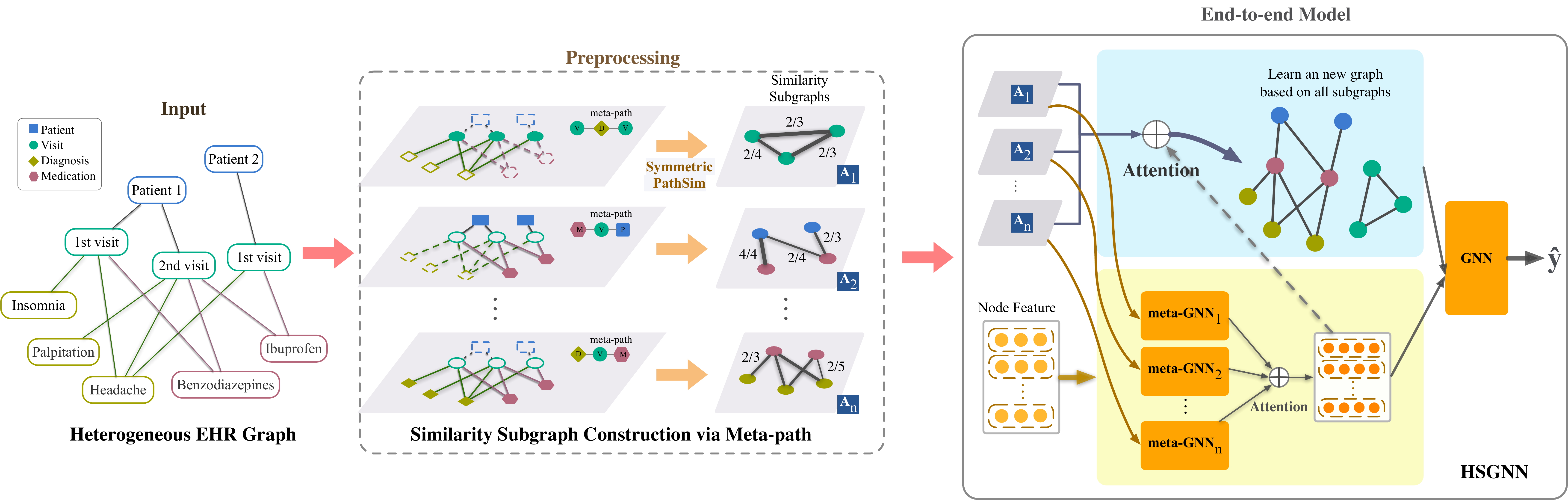} 
\caption{The proposed HSGNN framework. Heterogeneous EHR graph is preprocessed by calculating SPS along with each meta-path (the dash box) and then input into the end-to-end model (the solid box). Here we take meta-path V-D-V as an example to explain SPS. The 1st and 2nd visits of patient 1 have one common diagnosis in total, and therefore the numerator of similarity between them is 1*2=2. Besides, they have 4 diagnosis neighbors in total, and thus the denominator is 4. The similarity of these two nodes along with meta-path V-D-V is 1/2.}
\label{overview}
\end{figure*}

\section{Methods}

HSGNN consists of two parts: one is preprocessing step that splits the heterogeneous graph into multiple subgraphs; and the other is an end-to-end graph neural network that takes multiple graphs as input. In the first part, we introduce the definition of heterogeneous EHR graph, meta-path and symmetric PathSim (SPS). In the second part, we provide the forward propagation rules of our model. In this section we use the EHR with the same structure as Fig.\ref{hin} to introduce our model.

\begin{table}[tb]
  \centering
  \caption{Notations}
  \vspace{2mm}
    \resizebox{0.48\textwidth}{!}{\begin{tabular}{cl}
    \toprule
    Notation                    &Explanation \\
    \midrule
    
    $n_i$ & Node $i$ in the heterogeneous EHR graph.\\
    $\phi(\cdot)$ & Mapping function to retrieve the type of node. \\
    $PathCount_p(\cdot,\cdot)$ or $PC_p(\cdot,\cdot)$  & PathCount w.r.t meta-path $p$.\\
    $SPS_p(\cdot,\cdot)$ &Symmetric PathSim w.r.t meta-path $p$. \\
    $\bm A_{k}$ & The $k$-th input adjacency matrix.\\
    $\bm F$ & The input node features.\\
    $K$ & Number of meta-paths.\\
    $N$ & Number of nodes in the graph. \\
    $\bm A_{meta}$ & The fused adjacency matrix.\\
    $\bm F_{meta}$ & The aggregated node features.\\
    $\bm w$ or $\bm W$ & Parameters used to derive $\bm A_{meta}$.\\
    $\bm \Omega$ & Parameters used to derive $\bm W$.\\
    $meta\_GNN_{k}(\cdot,\cdot)$ & meta GNN module for the $k$-th meta-path.\\
    $AGGREGATOR_F(\cdot)$ & Aggregation function to derive $\bm F_{meta}$.\\
    \bottomrule
    \end{tabular}}
  \label{table:notation}
\end{table}

\subsection{Similarity Subgraph Construction via Meta-path}

Since the heterogeneous EHR graph consists of multiple types of node and edge, traditional GNN cannot process it directly. A approach is to process each node in the graph according to the node types\cite{DBLP:conf/kdd/ZhangSHSC19}. However, the links between different types of nodes can form some unique patterns and may possess specific meaning. Therefore, we introduce meta-path to process the heterogeneous graph and then calculate similarities between nodes along with each meta-path.

\paragraph{\textbf{Heterogeneous EHR Graph}} As shown in left part of Fig. \ref{hin}, a heterogeneous EHR graph consists of medical information from all patients. There are four kinds of nodes in the graph, patient $c$, visit $v$, diagnosis $d$ and medication $m$. Formally, we use $S = C+V+D+M$ to represent the set of all nodes in the graph, where $C$, $V$, $D$ and $M$ correspond to sets of patients, visits, diagnoses and medications. 
For each node $n \in S$, we also define a mapping $\phi(n) \in \{``C",``V",``D".``M"\}$ to find its type.

\paragraph{\textbf{Meta-path}} 
A meta-path $p = t_1 t_2\cdots t_n$ is a sequence where $ t \in \{``C",``V",``D".``M"\}$. It can represent a pattern of node types in a given path. For example, a meta-path $"VDV"$ denotes the pattern of ``visit node - diagnosis node - visit node" in the heterogeneous graph, and the path ``patient 1's 1st visit - headache - patient 2's 1st visit" is an instance of this meta-path.

\paragraph{\textbf{PathCount}}
Suppose we have two nodes $n_i,n_j \in S $ and a meta-path $p = t_1 t_2\cdots t_n$ where $\phi(n_i) = t_1$ and $\phi(n_j) = t_n$.
The $PathCount$ (shortened as $PC$) of $n_i,n_j$ w.r.t. $p$ is a function of the number of meta-path instances between node pairs.  
For example, in Fig. \ref{overview} the $PC$ under mata-path ``DVM" between node pair (``headache",``benzodiazepines") is 2, since they have 2 common visit neighbors.

\paragraph{\textbf{Symmetric PathSim} (SPS)}
Inspired by \cite{DBLP:journals/pvldb/SunHYYW11}, we propose the symmetric PathSim (SPS) used to measure the similarity of a node pair $(n_i,n_j)$ under a specific meta-path $p$ in the heterogeneous graph.
\begin{equation}
\label{pim}
SPS_p(n_i,n_j) =  \frac{PC_{p}(n_i,n_j)+PC_{p}(n_j,n_i)}{PC_{p}(n_i,n_i)+PC_{p}(n_j,n_j)} .
\end{equation}

Basically, when the $PC$ between two nodes is higher, these two nodes tend to have a stronger relation.
However, some nodes may have higher degree but are less important. For example, a node denoting gender ``female" may link to half of the patient nodes in the graph, but the effect of gender on medication is much less than the effect of diagnosis. To eliminate the influence of nodes with high visibility (degree) and low importance, SPS normalizes  the $PC$ with the sum of $n_i$ and $n_j$'s self loop count. SPS is symmetric, which means $SPS_p(n_i,n_j) = SPS_p(n_j,n_i)$.

In the preprocessing step, we construct the heterogeneous EHR graph and calculate the similarities of all node pairs under a group of meta-paths $P = \{p_1,p_2,\cdots\,p_{K}\}$ (the similarity of two nodes is set to 0 if their node types are not applicable to the mata-path). After this step, we can obtain a series of symmetric similarity matrices $\mathcal{A} = \{\bm A_{1}, \bm A_{2}, \cdots , \bm A_{K} \}$ where $K$ is both the number of meta-paths and the number of similarity matrices. The size of each matrix $\bm A_i$ in $\mathcal{A}$ is ${N \times N}$, where $N = |S|$ is the number of nodes. In this way, the heterogeneous graph is split into multiple homogeneous graphs and each homogeneous graph contains partial information of the original graph. 

\begin{figure*}[t]
\centering
\includegraphics[width=2.05\columnwidth]{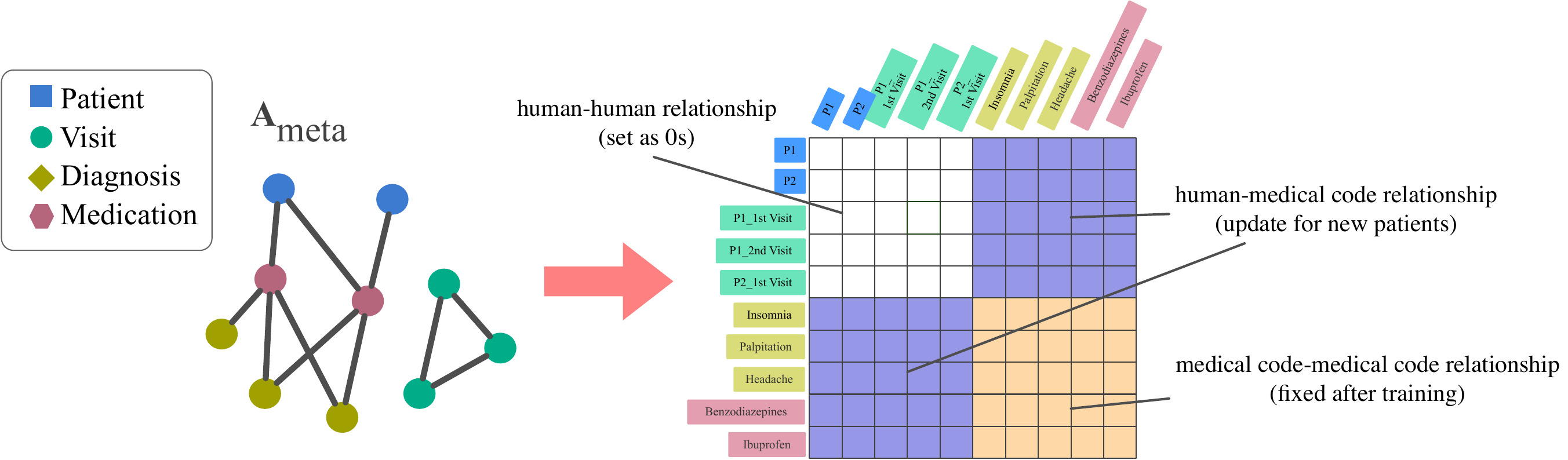} 
\caption{A dissection of $A_{meta}$.}
\label{retrain}
\end{figure*} 
\subsection{Heterogeneous Similarity Graph Neural Network}
The solid box in Fig. \ref{overview} shows the architecture of our proposed HSGNN. The preprocessing step derives multiple homogeneous graphs with meta-path and then we take them as inputs of HSGNN. The primary goal of HSGNN is to fuse the homogeneous graphs into one graph $A_{meta}$ containing true relations between each node pair. Then, $A_{meta}$ can be used for later general GNN layers such as Graph Convolutional Network (GCN) \cite{DBLP:conf/iclr/KipfW17} or for other downstream tasks. To achieve this goal, suppose the initial node feature matrix is $\bm F$ and the $K$ input graphs are $\mathcal{A} = \{\bm A_{1}, \bm A_{2}, \cdots , \bm A_{K} \}$, here we propose several variants of HSGNN.

\subsubsection{Simply Weighted Sum}
A straightforward approach is to use weighted sum:

\begin{equation}
\label{ws}
\bm A_{meta} = \sum_{k=1}^K  w_k \bm A_k
\end{equation}

where $w_k$ is a trainable scalar weight of matrix $\bm A_i$ and $\sum_{k=1}^K  w_k = 1$. An advantage of this approach is its simplicity. However, this method assumes the effect of one meta-path keeps constant over all nodes in the graph, regardless the uniqueness of each node. For example, to predict the condition of a patient, doctors may  rely on different medical codes when making decisions. 
Since medical codes correspond to different meta-paths, we need to adjust weight scalars on each node pair.

\subsubsection{Attention Sum}
We have a node feature matrix $\bm F$ as the input, and it can help us learn the proper weights of each graph. Since we want to assign a unique weight for each node pair under each meta-path, the weight tensor can be denoted as $\bm W \in [0,1]^{K \times N \times N}$ and each element $w_{kij}$ in it means the attention weight under node pair $(n_i, n_j)$ on the $k$-th meta-path. Similarly, we need to make sure $\sum_{k=1}^K  w_{kij} = 1$.

We adopt a one-layer feed forward neural network to calculate the attention value $w_{kij}$. The neural network takes two node features $\bm f_i$ and $ \bm f_j \in \bm F$ as the input, and outputs the weight of this node pair on all graphs. Formally, we have:

\begin{equation}
\label{att}
w_{k,i,j}= {\text{softmax}}_{k}(att_{ij})=\frac{\text{exp}(\sigma(\bm \omega_k^T  [\bm f_i ||\bm  f_j]))}{\sum_{l=1}^{K}\text{exp}(\sigma(\bm \omega_l^T  [\bm f_i ||\bm  f_j]))}.
\end{equation}

In Eq. (\ref{att}), $\bm  f_i$ and $\bm  f_j$ are feature vectors of node $n_i$ and $n_j$ and $||$ denotes concatenation operation. $\bm \Omega_{att} = \{\bm \omega_1; \bm \omega_2; \cdots ;\bm \omega_{K}\}$ is the parameter set of the neural network. After obtaining $w_{kij}$, we can get $A_{meta}$:

\begin{equation}
\label{ameta}
\bm A_{meta} = \sum_{k=1}^K  \bm W_k \circ \bm A_k
\end{equation}

where $\bm W_k$ means the $k$-th $N \times N$ matrix in $\bm W$ and $\circ$ means element-wise multiplication.

This equation adjusts personalized weights for different node pairs based on node features. However, this approach fails to improve the performance in the experiments. The reason is that, the node feature $\bm F$ we use in the experiments is not informative, and thus it can introduce noise into the model,  and prevent it from learning meaningful attention weights. To address this issue, we need to let the node features firstly learn from $\mathcal{A}$, and then use them to generate meaningful attention weights.

\subsubsection{Aggregated Attention Sum}

After learning from  $\mathcal{A}$ to obtain a more informative node feature matrix $\bm F_{meta}$, we use $\bm F_{meta}$ to generate the attention weights of graph aggregation. Motivated from \cite{DBLP:conf/www/WangJSWYCY19}, in this step we apply GNN on each graph to obtain multiple features for each node. Formally, for $k \in \{1,2,\cdots,K\}$ we have:

\begin{equation}
\label{fmeta0}
\bm F^{(0)}_{k} = meta\_GNN_{k} (\bm F, \bm A_k)
\end{equation}

where $meta\_GNN$ can be any kind of GNN layers. In the next step, to learn the node feature matrix $\bm F_{meta}$, we use

\begin{equation}
\label{fmeta}
\bm F_{meta} = AGGREGATOR_F([\bm F^{(0)}_{1};\bm F^{(0)}_{2};\cdots;\bm F^{(0)}_{K}]),
\end{equation}

where $AGGREGATOR_F$ is the aggregation function, which can be Graph Attention Network (GAT) \cite{DBLP:journals/corr/abs-1710-10903}. Here we can also use some other operations such as concatenate or average $\bm F^{(0)}_{1},\bm F^{(0)}_{2},\cdots,\bm F^{(0)}_{K}$ together.

When we get the $\bm F_{meta}$, we can use Eq. \ref{att} to learn the attention weights on graphs:
\begin{equation}
\begin{split}
\label{att2}
w_{k,i,j}= {\text{softmax}}_{k}(meta_att_{ij})=\\
\frac{\text{exp}(\sigma(\bm \omega_k^T  [\bm f^{meta}_i ||\bm  f^{meta}_j]))}{\sum_{l=1}^{K}\text{exp}(\sigma(\bm \omega_l^T  [\bm f^{meta}_i ||\bm  f^{meta}_j]))}.
\end{split}
\end{equation}

Many kinds of operations and aggregators can be used as $AGGREGATOR_F$. Here we provide several options which are compared in the experiments. Suppose previously we obtain $K$ node feature matrices $\bm F^{(0)}_1,\bm F^{(0)}_2,\cdots, \bm F^{(0)}_K $, we propose the following aggregation functions in our model.
\begin{itemize}
    \item Mean operation. That is,
    \begin{equation}
    \label{magg}
    \bm F_{meta} = \sum_{k=1}^K \bm F^{(0)}_k / K
    \end{equation}.
    
    \item Concatenation operation. That is,
    \begin{equation}
    \label{conagg}
    \bm F_{meta} = CONCAT([\bm F^{(0)}_1;\bm F^{(0)}_2;\cdots; \bm F^{(0)}_K])
    \end{equation}.
    
\end{itemize}

After obtaining $\bm F_{meta}$ and $\bm A_{meta}$, we use general GNN layers such as GCN \cite{DBLP:conf/iclr/KipfW17} and GAT \cite{DBLP:journals/corr/abs-1710-10903} to derive final predictions.

\subsection{Quick Inference When New Records Coming}

Basically, HSGNN needs all nodes in the graph to present during training and thus is transductive. According to \cite{hamilton2017inductive}, transductive GNN cannot handle new nodes and edges without re-training. However, there is a special characteristic of EHR graph: the number of all medical code nodes, such as diagnosis node and medication node, keep almost constant in all EHR graphs. The total number of all diagnoses, medications, procedures and lab tests in real-world dataset is about 5000 and they seldom change. This number is relatively small and their similarities can be easily stored in the memory. Another fact is that new coming patients/visits are never isolated, as they always appear with some medical features. In other words, there are always ``patient/visit-medical code" links in the test set. Therefore, using these two properties, we can use HSGNN to infer new patients/visits without re-train the model.

After the training step, we obtain a well-trained $\bm A_{meta}$ in HSGNN. Since  $\bm A_{meta}$ contains medical relations, it can be used in the inference step. As shown in Fig. \ref{retrain}, when we dissect $\bm A_{meta}$, all edges in $\bm A_{meta}$ can be grouped into three categories. 
\begin{itemize}
    \item Medical code-medical code edges. Edges between two medical codes such as ``diagnosis-medication" relation reveals the relationship between medical factors. Weights of these edges keep stable after training and can be reused in the inference step.
    \item Human-medical code edges. These edges represent the relationship between a human (patient/visit node) and a medical code. Since human nodes are different in training and testing step, weights of these edges cannot be reused. However, we can calculate these weights in the preprocessing step using testing data under ``human - $\cdots$ - medical code" meta-paths.
    \item Human-human edges. Weights in this part is set to 0s since there is no way to calculate them. The volume of testing data is relatively small and we still have other edges available, so these 0s won't interfere prediction.
\end{itemize}

After obtaining a new $\bm A_{meta}$ for testing set, we can use general GNNs to predict testing results. More details about this part will be provided in the experiment section.

\begin{figure}[t]
\centering
\includegraphics[width=0.8\columnwidth]{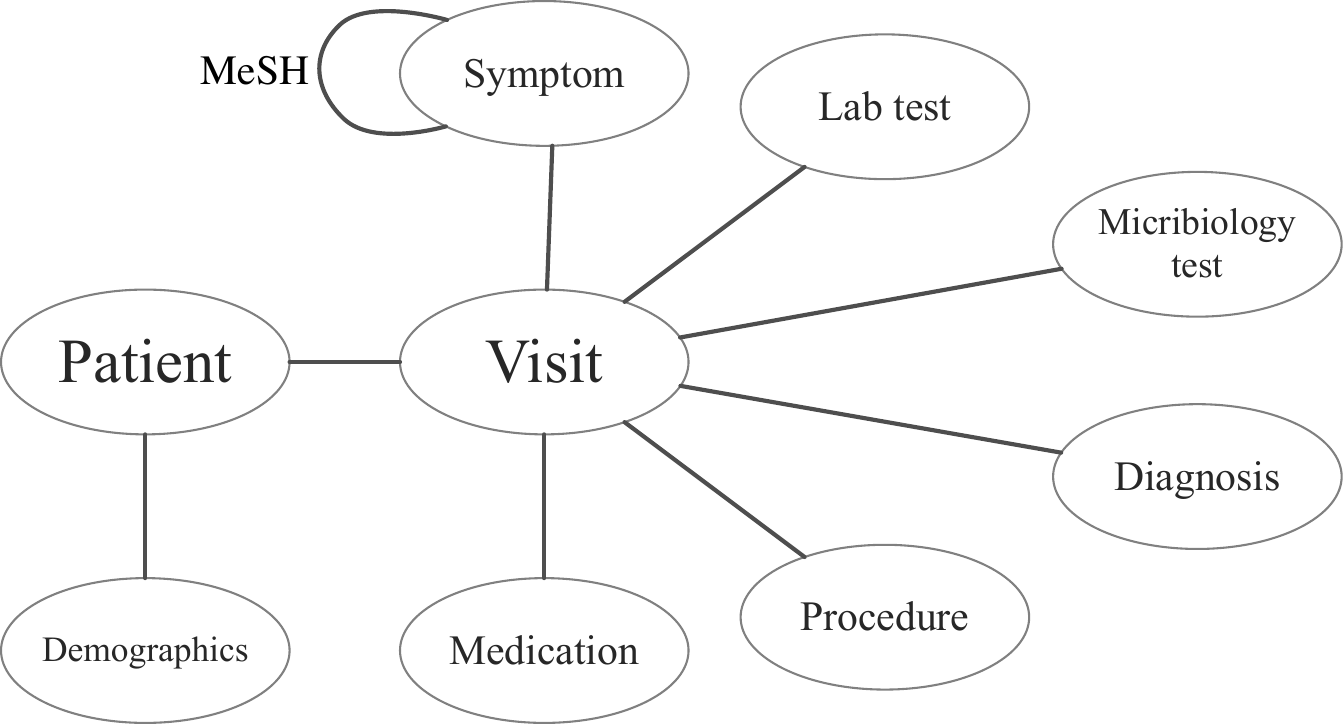} 
\caption{The data schema of the MIMIC-III network.}
\label{schema}
\end{figure}

\section{Experiments}

In this section, we conduct experiments on the public MIMIC dataset and show the superiority of HSGNN over other baselines. 

\subsection{MIMIC-III Dataset}

MIMIC \cite{goldberger2000physiobank,johnson2016mimic} is a publicly available dataset consisting of health records of 46,520 intensive care unit (ICU) patients over 11 years. Table \ref{statistics} shows the statistics of the graph we construct and Fig. \ref{schema} shows the structure of MIMIC-III dataset.

Raw MIMIC-III data consists of 17 tables, including demographics, laboratory test results, microbiology test results, diagnoses, medications, procedures, medical notes, etc. For each patient and visit, there is a unique ID to track its corresponding information through tables. There are extra tables recording the patient-visit relations, demographics and data dictionaries as well. To build a clean and efficient heterogeneous graph based on these data, we mainly do the following things.
\begin{table}[]
\centering
\caption{ Node statistics for the MIMIC-III network.}
\label{statistics}
\begin{tabular}{@{}ccc@{}}
\toprule
MIMIC-III & \# of code & \begin{tabular}[c]{@{}c@{}}Avg \# of the code \\ (for each visit)\end{tabular} \\ \midrule
Patient & 46,520 & -- \\
Visit & 58,976 & -- \\
Diagnosis & 203 & 11.20 \\
Procedure & 157 & 4.65 \\
Medication & 304 & 23.18 \\
Lab tests & 480 & 27.55 \\
Microbiology tests & 258 & 0.94 \\
Symptoms & 324 & 19.06 \\ \bottomrule
\end{tabular}
\end{table}

\begin{table*}[]
\centering
\caption{Overall top-$k$ precision of all baselines and HSGNN variants on MIMIC-III dataset.}
\label{overall}
\resizebox{0.8\textwidth}{!}{%
\begin{tabular}{@{}lcccccccc@{}}
\toprule
\multicolumn{1}{c}{\multirow{2}{*}{Model}} & \multicolumn{4}{c}{Visit-level precision@k} & \multicolumn{4}{c}{Patient-level precision@k} \\ \cmidrule(l){2-9} 
\multicolumn{1}{c}{} & 5 & 10 & 15 & \multicolumn{1}{c|}{20} & 5 & 10 & 15 & 20 \\ \midrule
\multicolumn{1}{l|}{Dipole} & 0.5929 & 0.7426 & 0.7942 & \multicolumn{1}{c|}{0.7540} & 0.6393 & 0.7359 & 0.7271 & 0.7239 \\
\multicolumn{1}{l|}{KAME} & 0.6107 & 0.7475 & 0.7967 & \multicolumn{1}{c|}{0.7562} & 0.6472 & 0.7565 & 0.7530 & 0.7288 \\
\multicolumn{1}{l|}{HeteroMed} & 0.5893 & 0.7314 & 0.7866 & \multicolumn{1}{c|}{0.7670} & 0.6285 & 0.7255 & 0.7171 & 0.7193 \\
\multicolumn{1}{l|}{MAGNN} & 0.6142 & 0.7471 & 0.8092 & \multicolumn{1}{c|}{0.7693} & 0.6501 & 0.7585 & 0.7548 & 0.7394 \\
\multicolumn{1}{l|}{HAN} & 0.6135 & 0.7464 & 0.8083 & \multicolumn{1}{c|}{0.7691} & 0.6494 & 0.7582 & 0.7550 & 0.7386 \\
\multicolumn{1}{l|}{HetGNN} & 0.6124 & 0.7456 & 0.8070 & \multicolumn{1}{c|}{0.7689} & 0.6489 & 0.7580 & 0.7452 & 0.7379 \\
\multicolumn{1}{l|}{GCT} & 0.6297 & 0.7503 & 0.8107 & \multicolumn{1}{c|}{0.7703} & 0.6633 & 0.7592 & 0.7685 & 0.7384 \\ \midrule
\multicolumn{1}{l|}{HSGNN} & \textbf{0.6426} & \textbf{0.7658} & \textbf{0.8189} & \multicolumn{1}{c|}{\textbf{0.7736}} & \textbf{0.6778} & 0.7613 & 0.7702 & \textbf{0.7456} \\
\multicolumn{1}{l|}{simi-HSGNN} & 0.6123 & 0.7396 & 0.8034 & \multicolumn{1}{c|}{0.7689} & 0.6488 & 0.7481 & 0.7452 & 0.7479 \\
\multicolumn{1}{l|}{sum-HSGNN} & 0.6412 & 0.7630 & 0.8129 & \multicolumn{1}{c|}{0.7683} & 0.6724 & 0.7597 & 0.7696 & 0.7384 \\
\multicolumn{1}{l|}{HSGNN-m} & 0.6410 & 0.7638 & 0.8175 & \multicolumn{1}{c|}{0.7667} & 0.6752 & \textbf{0.7740} & \textbf{0.7723} & 0.7429 \\ 


\bottomrule
\end{tabular}%
}
\end{table*}

\paragraph{data disambiguation}
There are more than 1000 kinds of medications in the original dataset. Most of them are different abbreviations or preparations of the same medicine. In the experiment, we disambiguate these medicines by comparing the most common strings in the name of medications and finally extract 304 most common medications.

\paragraph{continuous variables bucketization}
Lab test results are mostly continuous values. Therefore we need to bucketize them into discrete variables and integrate these variables into the graph. Some of the entries in the lab test table in MIMIC-III contain a ``N/A" flag, indicating whether the test result is normal or abnormal. We then set up two nodes for this lab test representing  ``normal" and ``abnormal". For other lab tests that do not have such an flag, we use the quartiles of the lab test to bucketize the outliers from common values. Some data engineering works are also conducted in this step to make sure we get sensible thresholds.

\paragraph{medical notes preprocessing}
There are no symptom records for patients in MIMIC-III, but there exist medical nodes for each visit. Medical notes contain rich diagnostic information but are difficult to process since they are free texts. To extract diagnostic information (symptoms) from them without data leakage, we use an extra knowledge graph MeSH (Medical Subject Headings)\footnote{\url{https://meshb.nlm.nih.gov/treeView}} to extract meaningful structural diagnostic information from free text. We extract entries ``medications on admission", ``family history", ``impression", ``chief complaint", ``physical examination on admission" and ``history" from the medical notes, and then match words in these entries to MeSH. After that we use these matched keyworks together with its connections in MeSH to help building the heterogeneous MIMIC-III network. By doing this we extract keywords from the diagnostic texts while incorporating external knowledge graph into our graph.

\paragraph{other medical codes}
MIMIC-III uses ICD-9-PC and ICD-9 ontology to represent all procedures and diagnoses. International Classification of Diseases (ICD)  is a medical ontology which is widely used in healthcare. In these ontologies, diagnoses and procedures are organized in hierarchical structures and the first several digits denote a high-level concept of the codes. In this case, we choose the first two digits for procedure codes and the first three digits for diagnosis codes to predict. We then select the most commonly existing codes as nodes in our graph and dismiss other rare codes.

\begin{figure*}
        \centering
        \begin{subfigure}[b]{0.24\textwidth}
            \centering
            \includegraphics[width=1.65in]{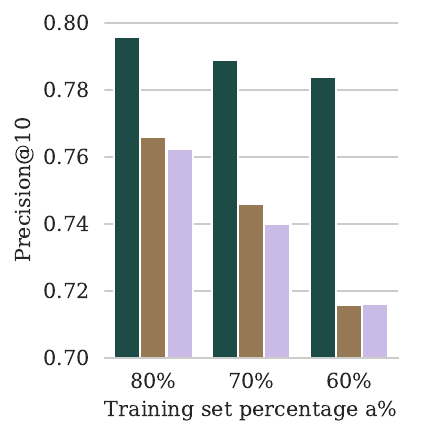}
            {{\small Visit-level precision.}}    
            \label{fig:mean and std of net14}
        \end{subfigure}
        \hfill
        \begin{subfigure}[b]{0.24\textwidth}  
            \centering 
            \includegraphics[width=1.65in]{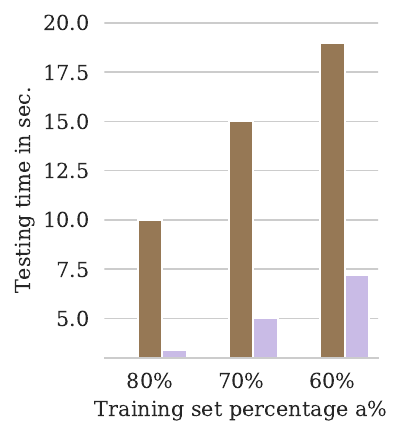}
            {{\small Visit-level time.}}    
            \label{fig:mean and std of net24}
        \end{subfigure}
        \hfill
        \begin{subfigure}[b]{0.24\textwidth}   
            \centering 
            \includegraphics[width=1.65in]{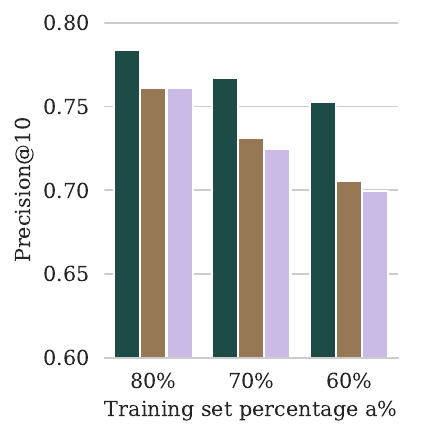}	
            {{\small Patient-level precision.}}    
            \label{fig:mean and std of net34}
        \end{subfigure}
        \hfill
        \begin{subfigure}[b]{0.24\textwidth}   
            \centering 
            \includegraphics[width=1.65in]{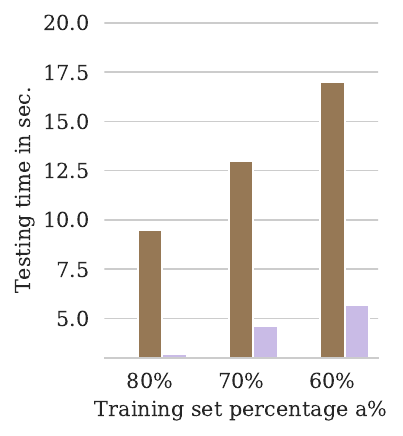}	
            {{\small Patient-level time.}}    
            \label{fig:mean and std of net44}
        \end{subfigure}
      
        {\small Fig. 5: Precision and running time of Quick Inference compare with traditional train and test procedure. 
        (Dark green denotes training, brown denotes testing and purple denoted quick reference.)} 
\label{time}
\end{figure*}

\subsection{Baselines}
To demonstrate the advantage of HSGNN, we select three medical predictive models and three graph neural networks as our baseline.

\begin{itemize}[noitemsep,topsep=0pt]
    \item\textbf{Dipole} \cite{DBLP:conf/kdd/MaCZYSG17}. Dipole uses bidirectional recurrent neural networks and attention mechanism to make predictions. In this experiment, we use patient conditions at different times in one visit to make the visit-level prediction, and use information of different visits to make patient-level prediction.
    
    \item\textbf{KAME} \cite{DBLP:conf/cikm/MaYXCZG18}. KAME learns to predict patients’ health situation. It incorporates medical knowledge graph, and utilizes attention mechanism to make accurate predictions. We leverage the MeSH ontology as the knowledge graph to run this model. It is fair to compare KAME and our proposed HSGNN  becuase they all make use of the same ontology although in different settings.

    \item\textbf{HeteroMed} \cite{DBLP:conf/cikm/HosseiniCWSS18}. HeteroMed is the first approach using HIN to process EHR. It exploits meta-paths and employs a joint embedding framework to predict diagnosis for patients. We use the same graph structure on this model as HSGNN.
    
    \item\textbf{MAGNN}
    \cite{DBLP:conf/www/0004ZMK20}. MAGNN proposes intra-metapath aggregators and inter-metapath aggregators to make inductive predictionson heterogeneous graphs. We use the same graph structure and meta-paths on this model as HSGNN.
     
    \item\textbf{HetGNN}
    \cite{DBLP:conf/kdd/ZhangSHSC19}.
    HetGNN is a heterogeneous graph neural network that introduces a random walk to sample a fixed size of heterogeneous neighbors and leverages a neural network architecture with two modules to aggregate feature information of those sampled neighboring nodes

    \item\textbf{HAN}
    \cite{DBLP:conf/www/WangJSWYCY19}.
    HAN 
    is a heterogeneous graph neural network based on hierarchical attention, including node-level and semantic-level attentions to learn the importance between a node and its metapath based neighbors and the importance of different meta-paths.
    
    \item\textbf{GCT}
    \cite{choilearning}. GCT uses graph convolutional transformers to jointly learn the hidden structure of EHR while performing prediction tasks on EHR data. GCT uses data statistics to guide the structure learning process.
    In the experiments, we use the data schema mentioned above to generate its pre-training weights.

\end{itemize}

Meanwhile, we also conduct experiments on following five variants on HSGNN to find the best architecture.

\begin{itemize}[noitemsep,topsep=0pt]
    \item \textbf{HSGNN} Model we proposed in this paper, using concatenation operation to derive $\bm F_{meta}$  (Eq. \ref{fmeta0} \ref{conagg}) and aggregated attentional sum to derive $\bm A_{meta}$ (Eq. \ref{att2} \ref{ameta}). Then a one-layer GCN is applied on $\bm F_{meta}$ and $\bm A_{meta}$ to make final predictions.
    
    \item\textbf{simi-HSGNN} Use $PathCount$ but not SPS to derive $\mathcal{A}$. This is to show the efficiency of SPS.
    
    \item\textbf{sum-HSGNN}
    Use simply weighted sum to derive $\bm A_{mate}$ (Eq. \ref{ws}). Then a one-layer GCN is applied on $\bm F$ and $\bm A_{meta}$ to make final predictions. This is to compare HSGNN with a simpler model to show the efficiency of splitting EHR graph into multiple subgraphs.
    
    \item\textbf{HSGNN-m}
    Use mean aggregator to derive $\bm A_{mate}$. Other settings are the same as HSGNN. This variant is to show the effect of different aggregation functions.

\end{itemize}

\subsection{Problem Introduction}
Diagnosis prediction can be viewed as a multi-label classification problem where we try to predict multiple possible diagnoses for the patients or visits. We conduct both patient level prediction and visit level prediction on the dataset. As for patient level prediction, only diagnoses existing on all visits of this patient would be counted as the diangosis of the patient. We then split training and testing set by removing the corresponding ``visit-diagnosis " edges in the graph. Then, since medication and procedure can be determined by diagnosis, these edges are also removed to prevent data leakage.

\subsection{Experiment Settings}
In the experiment, we use the concatenation of feature vectors from different sources as the features of the visits, and then we use them for all baseline models. For each experiment, we use 10-fold cross validation. Training, validation and testing sets are with a 7 : 1 : 2 ratio. Our method is implemented by Tensorflow 2.0 and Python 3.6, and tested on an machine with 32G RAM and 2 NVIDIA GeForce RTX 2080 Ti GPU. To evaluate the quality of prediction, we use precision at top-$k$ as the metric. We set the value of $k$ as 5, 10, 15, 20.

\subsection{Results of diagnosis prediction}

\begin{figure*}
        \centering
        \begin{subfigure}[b]{0.24\textwidth}
            \centering
            \includegraphics[width=1.65in,trim=43 25 43 40,clip]{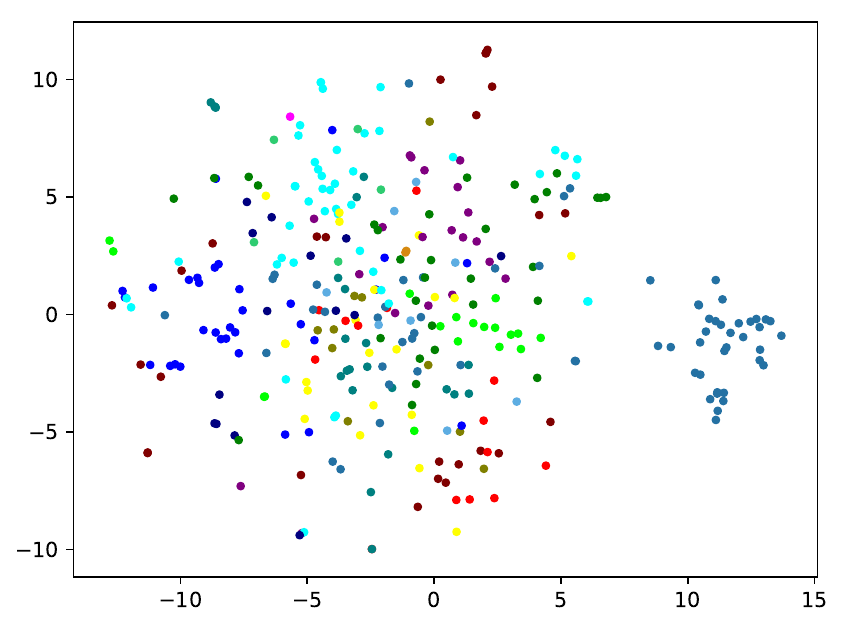}
         
            {{\small DeepWalk.}}    
            \label{fig:mean and std of net14}
        \end{subfigure}
        \hfill
        \begin{subfigure}[b]{0.24\textwidth}  
            \centering 
            \includegraphics[width=1.65in,trim=43 25 43 40,clip]{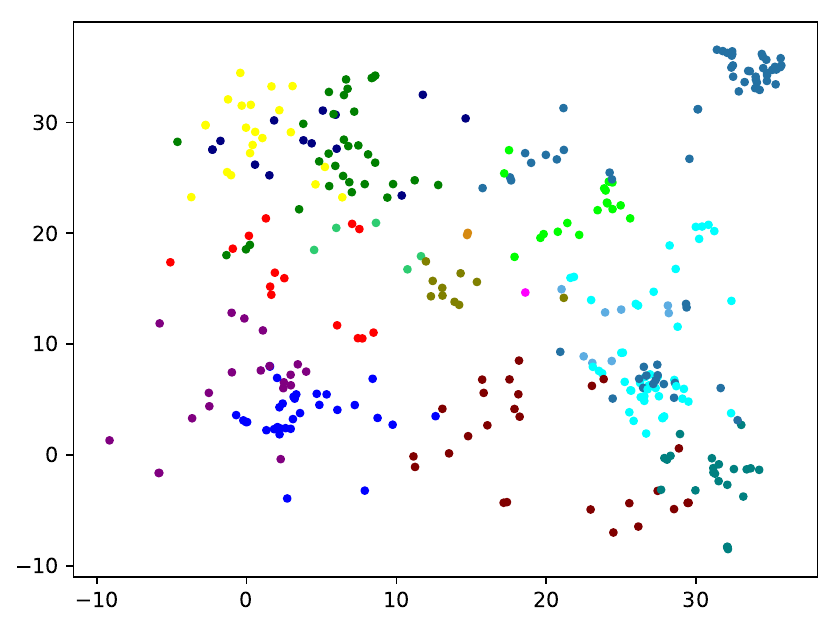}
           
            {{\small metapath2vec.}}    
            \label{fig:mean and std of net24}
        \end{subfigure}
        \hfill
        \begin{subfigure}[b]{0.24\textwidth}   
            \centering 
            \includegraphics[width=1.65in,trim=43 25 43 40,clip]{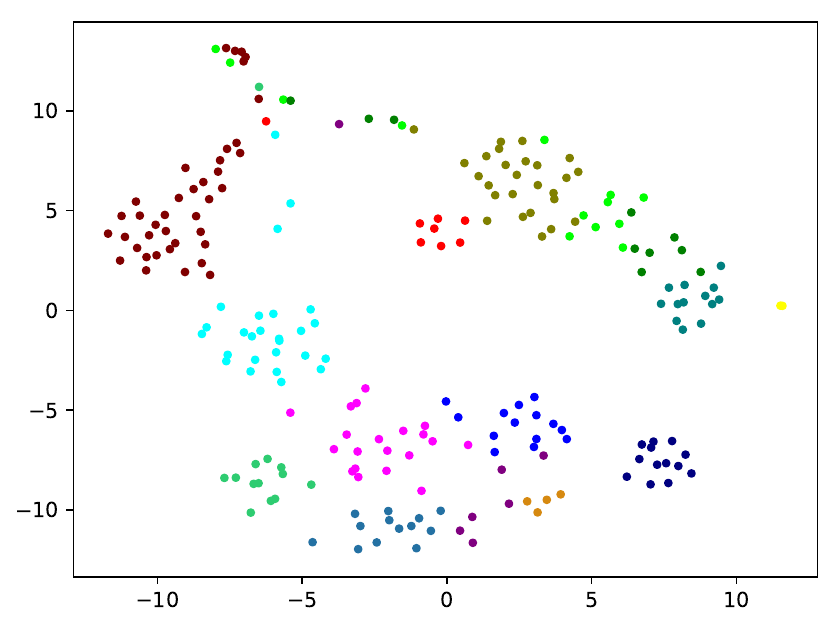}	
          
            {{\small GRAM.}}    
            \label{fig:mean and std of net34}
        \end{subfigure}
        \hfill
        \begin{subfigure}[b]{0.24\textwidth}   
            \centering 
            \includegraphics[width=1.65in,trim=45 25 15 15,,clip]{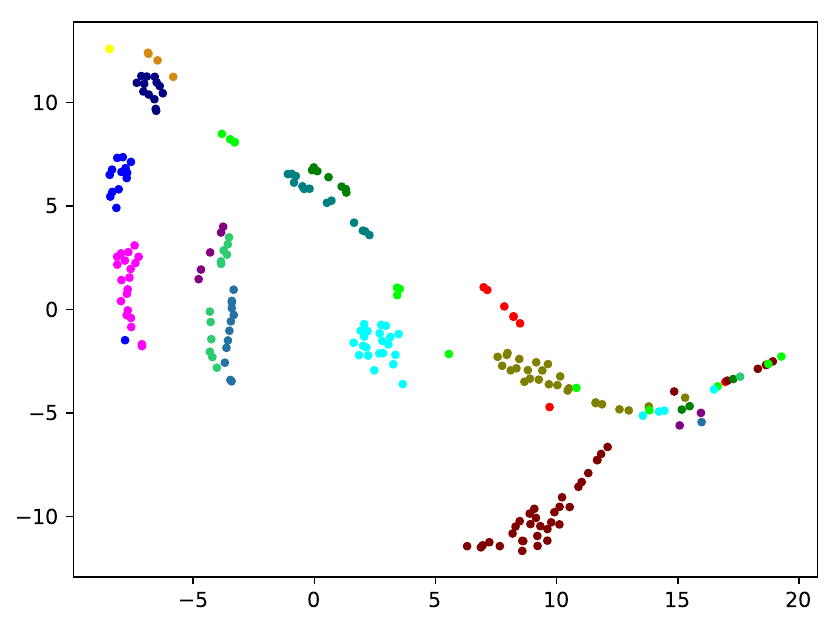}	
          
            {{\small HSGNN.}}    
            \label{fig:mean and std of net44}
        \end{subfigure}
      
        {\small Fig. 6: T-SNE scatterplots of diagnoses trained by HSGNN, DeepWalk, metapath2vec and GRAM.} 
\label{vis}
\end{figure*}    

\subsubsection{Comparison with other baselines}
Table \ref{overall} displays the performance of all comparable models on MIMIC-III. In the table, HSGNN and its variant HSGNN-m. outperform all other baselines. 
We conduct diagnosis prediction task the MIMIC-III dataset. Generally, there are about 10 diagnoses for each visit and 4 visits for each patient. Therefore, when $k$ increases, the precision may either increase or decrease. The accuracy of a model approximately reach its maximum when $k=10$ for patient diagnosis prediction and $k=15$ for visit level prediction. This is also why we choose maximum $k=20$. Therefore, if we focus on the column of $k=15$ of the visit-level prediction and $k=10$ of patient-level prediction, we can find HSGNN improve 0.7\% and 1.4\% on both tasks. 

All baselines, together with HSGNN can be classified into three categories: RNN models, homogeneous graph models and heterogeneous graph models. From the results we can infer that homogeneous graph models (KAME and GCT) perform better than RNN models (Dipole), and heterogeneous graph approaches (MAGNN and HSGNN) perform better than homogeneous approaches. It demonstrates the effectiveness of considering structural information when making predictions. Compared with homogeneous graphs, heterogeneous graphs carry more information and thus can achieve more improvement when applied to the model.

Among all baseline models, GCT achieves the best performance even if it uses homogeneous graph. Note that a common design of GCT and HSGNN is that they both use trainable weights and construct a virtual graph in the model. Therefore, we can infer that compared with using the original input graph, a virtual graph constructed in the model can improve the performance of GNN. Since our proposed HSGNN outperforms GCT, our model can learn a more accurate graph structure in the model. This phenomenon is because our model uses the heterogeneous graph as input and considers the difference between meta-paths.

\subsubsection{Comparison among HSGNN variants}
We also test some variants of HSGNN to find the best architecture of HSGNN while making some ablation studies. The first variant we compare with is simi-HSGNN, which uses the $PathCount$ as similarity measure but not SPS. By doing so, HSGNN becomes almost equivalent to HAN \cite{DBLP:conf/www/WangJSWYCY19} and its performance can be viewed as the performance of HAN. Simi-HSGNN performs worse than HSGNN for around 2\% on both tasks, showing that using normalize similarity measure SPS is an essential way to achieve better results.

Another variant considered is sum-HSGNN. Compared with HSGNN, sum-HSGNN is its simplified version since it contains less parameters in the model and is faster to train. Nevertheless, the performance of sum-HSGNN doesn't decrease a lot because of its simplicity and sum-HSGNN outperforms all other variants and baseline models except HSGNN and HSGNN-m. The reason may be that sum-HSGNN still preserves the mechanism of learning a trainable virtual graph.

HSGNN-mm shows the impact of different node aggregators on the model performance. However, we discover the influence of aggregators is limited if the size of embeddings are kept constant. Therefore, we choose the mean aggregator, the one which is easier to implement and can achieve satisfactory performance, to be compared in the experiments.


\subsection{Performance of Quick Inference}
To compare the efficiency and the effectiveness of our quick inference method (III. C) to traditional testing step, we design the following experiment to evaluate its performance. Firstly, we choose $a\%$ of data randomly from the dataset as training and validation samples. Then we split the remaining $1-a\%$ samples equally for traditional testing and quick inference. Secondly, in the preprocessing step, both training samples and testing samples are used to generate the graph. Then this graph is fed forward to our model. Finally, when the model is well-trained, we use the quick inference method to predict the remaining $(1-a\%)/2$ samples, and compare its precision and running time to the traditional testing procedure. In this experiment, we set $a=80\%,70\%,60\%$ respectively.

Fig. 5 shows the result of training performance, testing performance and the quick inference performance under visit-level and patient-level prediction. For each task, we evaluate the precision@10 of training samples, testing samples and quick inference samples after the model is well-trained. We also measure the time for testing samples and quick reference samples to get the results. We do not measure the time of training procedure because it depends on parameters such as learning rate. 

From Fig. 5 we can discover that the quick inference accuracy is only slightly lower then the traditional testing precision on both visit prediction and patient prediction tasks. Nevertheless, the time of getting quick inference results is much shorter than getting a traditional testing result. This is because quick inference can get $\bm A_{meta}$ without forward-propagation, and then get results simply through a one-layer graph neural network.

With $a\%$ decreasing, all the training, testing and quick inference precision decreases. It is because of the lack of training samples, making the model under-fitting. On the other hand, the decrease of training samples means the number of testing samples and quick inferences are increasing. Therefore, the number of inference is increasing.

\subsection{Representation Learning with External Knowledge}

HSGNN can learn representations for nodes. Since many models such as GRAM can learn high quality representations by integrating medical ontologies, we try to test the ability of HSGNN to learn informative representations on the same task.
In this experiment, we apply ICD-9 ontology on both GRAM and HSGNN to let them learn representations for diagnoses. Here are we choose nine categories in ICD-9 ontology to build the graph. Since diagnoses in the same category are directly connected and are more relative to each other, an ideal result is that all diagnosis nodes belong to the same category can form a cluster in visualization. To train HSGNN in an unsupervised way, we apply a loss like \cite{hamilton2017inductive} which maximizes the dot product of diagnoses in the same category. Fig. 6 shows the result of representation learning by ploting the t-SNE result \cite{DBLP:conf/vissym/RauberFT16}. Here we compare the results of HSGNN with GRAM, DeepWalk \cite{DBLP:conf/kdd/PerozziAS14} and metapath2vec \cite{DBLP:conf/kdd/DongCS17}. In Fig. 6, the colors of the dots represents the ICD-9 categories. According to the visualization, we can prove that HSGNN can produce representations with high quality since it forms clear clusters for each category.

\section{Conclusion}
EHR data is highly heterogeneous with high-dimensional temporal data. To model the intrinsic complexity of EHRs and utilize external medical knowledge, we propose HSGNN framework to learn high quality representations while generating predictions. HSGNN accepts similarity matrices as inputs and use the attention mechanism to measure the impact of each meta-paths. In the experiment section, we conduct diagnosis prediction task on MIMIC-III dataset, proving the superiority ability of HSGNN over baseline models. The visualization of representations shows the ability of HSGNN in generating reasonable representations both for diagnosis and patients. The superiority of HSGNN is mainly because it can make use of external medical ontologies together with both temporal and structural information.

\section*{Acknowledgment}
\addcontentsline{toc}{section}{Acknowledgment}
The corresponding author is Hao Peng. 
This work is supported by Key Research and Development Project of Hebei Province (No. 20310101D), NSFC No.62002007 and No.62073012, and in part by NSF under grants III-1763325, III-1909323, IIS-1763365 and SaTC-1930941.

\bibliographystyle{IEEEtran}
\bibliography{reference}

\begin{thebibliography}{10}
\providecommand{\url}[1]{#1}
\csname url@samestyle\endcsname
\providecommand{\newblock}{\relax}
\providecommand{\bibinfo}[2]{#2}
\providecommand{\BIBentrySTDinterwordspacing}{\spaceskip=0pt\relax}
\providecommand{\BIBentryALTinterwordstretchfactor}{4}
\providecommand{\BIBentryALTinterwordspacing}{\spaceskip=\fontdimen2\font plus
\BIBentryALTinterwordstretchfactor\fontdimen3\font minus
  \fontdimen4\font\relax}
\providecommand{\BIBforeignlanguage}[2]{{%
\expandafter\ifx\csname l@#1\endcsname\relax
\typeout{** WARNING: IEEEtran.bst: No hyphenation pattern has been}%
\typeout{** loaded for the language `#1'. Using the pattern for}%
\typeout{** the default language instead.}%
\else
\language=\csname l@#1\endcsname
\fi
#2}}
\providecommand{\BIBdecl}{\relax}
\BIBdecl

\bibitem{DBLP:conf/ijcai/FuHXS19}
T.~Fu, T.~N. Hoang, C.~Xiao, and J.~Sun, ``{DDL:} deep dictionary learning for
  predictive phenotyping,'' in \emph{IJCAI}, 2019, pp. 5857--5863.

\bibitem{bai2018ehr}
T.~Bai, A.~K. Chanda, B.~L. Egleston, and S.~Vucetic, ``Ehr phenotyping via
  jointly embedding medical concepts and words into a unified vector space,''
  \emph{BMC medical informatics and decision making}, vol.~18, no.~4, p. 123,
  2018.

\bibitem{DBLP:conf/kdd/ZhangTDZW19}
X.~S. Zhang, F.~Tang, H.~H. Dodge, J.~Zhou, and F.~Wang, ``Metapred:
  Meta-learning for clinical risk prediction with limited patient electronic
  health records,'' in \emph{SIGKDD}, 2019, pp. 2487--2495.

\bibitem{lipton2015learning}
Z.~C. Lipton, D.~C. Kale, C.~Elkan, and R.~Wetzel, ``Learning to diagnose with
  lstm recurrent neural networks,'' \emph{arXiv preprint arXiv:1511.03677},
  2015.

\bibitem{DBLP:conf/aaai/ShangXMLS19}
J.~Shang, C.~Xiao, T.~Ma, H.~Li, and J.~Sun, ``Gamenet: Graph augmented memory
  networks for recommending medication combination,'' in \emph{AAAI}, 2019, pp.
  1126--1133.

\bibitem{DBLP:conf/icdm/CheL17}
\BIBentryALTinterwordspacing
Z.~Che and Y.~Liu, ``Deep learning solutions to computational phenotyping in
  health care,'' in \emph{ICDM Workshops}.\hskip 1em plus 0.5em minus
  0.4em\relax {IEEE} Computer Society, 2017, pp. 1100--1109. [Online].
  Available: \url{https://doi.org/10.1109/ICDMW.2017.156}
\BIBentrySTDinterwordspacing

\bibitem{cai2018medical}
X.~Cai, J.~Gao, K.~Y. Ngiam, B.~C. Ooi, Y.~Zhang, and X.~Yuan, ``Medical
  concept embedding with time-aware attention,'' in \emph{Proceedings of the
  27th International Joint Conference on Artificial Intelligence}, 2018, pp.
  3984--3990.

\bibitem{choi2016multi}
E.~Choi, M.~T. Bahadori, E.~Searles, C.~Coffey, M.~Thompson, J.~Bost,
  J.~Tejedor-Sojo, and J.~Sun, ``Multi-layer representation learning for
  medical concepts,'' in \emph{SIGKDD}, 2016, pp. 1495--1504.

\bibitem{DBLP:conf/nips/ChoiBSKSS16}
E.~Choi, M.~T. Bahadori, J.~Sun, J.~Kulas, A.~Schuetz, and W.~F. Stewart,
  ``{RETAIN:} an interpretable predictive model for healthcare using reverse
  time attention mechanism,'' in \emph{NeurIPS}, 2016, pp. 3504--3512.

\bibitem{DBLP:conf/kdd/MaCZYSG17}
F.~Ma, R.~Chitta, J.~Zhou, Q.~You, T.~Sun, and J.~Gao, ``Dipole: Diagnosis
  prediction in healthcare via attention-based bidirectional recurrent neural
  networks,'' in \emph{SIGKDD}, 2017, pp. 1903--1911.

\bibitem{DBLP:journals/corr/LiptonKEW15}
Z.~C. Lipton, D.~C. Kale, C.~Elkan, and R.~C. Wetzel, ``Learning to diagnose
  with {LSTM} recurrent neural networks,'' in \emph{4th International
  Conference on Learning Representations, {ICLR} 2016}, 2016.

\bibitem{DBLP:conf/mlhc/ChoiBSSS16}
E.~Choi, M.~T. Bahadori, A.~Schuetz, W.~F. Stewart, and J.~Sun, ``Doctor {AI:}
  predicting clinical events via recurrent neural networks,'' in
  \emph{Proceedings of the 1st Machine Learning in Health Care, {MLHC} 2016},
  vol.~56.\hskip 1em plus 0.5em minus 0.4em\relax JMLR.org, 2016, pp. 301--318.

\bibitem{DBLP:journals/corr/AczonLHGFWW17}
M.~Aczon, D.~Ledbetter, L.~V. Ho, A.~M. Gunny, A.~Flynn, J.~Williams, and R.~C.
  Wetzel, ``Dynamic mortality risk predictions in pediatric critical care using
  recurrent neural networks,'' \emph{CoRR}, vol. abs/1701.06675, 2017.

\bibitem{DBLP:conf/cikm/MaYXCZG18}
F.~Ma, Q.~You, H.~Xiao, R.~Chitta, J.~Zhou, and J.~Gao, ``{KAME:}
  knowledge-based attention model for diagnosis prediction in healthcare,'' in
  \emph{CIKM}, 2018, pp. 743--752.

\bibitem{choilearning}
E.~Choi, Z.~Xu, Y.~Li, M.~W. Dusenberry, G.~Flores, E.~Xue, and A.~M. Dai,
  ``Learning the graphical structure of electronic health records with graph
  convolutional transformer,'' in \emph{Proceedings of the Thirty-Second {AAAI}
  Conference on Artificial Intelligence}.\hskip 1em plus 0.5em minus
  0.4em\relax {AAAI} Press, 2020.

\bibitem{DBLP:conf/kdd/ChoiBSSS17}
E.~Choi, M.~T. Bahadori, L.~Song, W.~F. Stewart, and J.~Sun, ``{GRAM:}
  graph-based attention model for healthcare representation learning,'' in
  \emph{KDD}, 2017, pp. 787--795.

\bibitem{DBLP:conf/nips/ChoiXSS18}
E.~Choi, C.~Xiao, W.~F. Stewart, and J.~Sun, ``Mime: Multilevel medical
  embedding of electronic health records for predictive healthcare,'' in
  \emph{NIPS 2018}, 2018, pp. 4552--4562.

\bibitem{DBLP:conf/www/WangJSWYCY19}
X.~Wang, H.~Ji, C.~Shi, B.~Wang, Y.~Ye, P.~Cui, and P.~S. Yu, ``Heterogeneous
  graph attention network,'' in \emph{The World Wide Web Conference, {WWW}
  2019}.\hskip 1em plus 0.5em minus 0.4em\relax {ACM}, 2019, pp. 2022--2032.

\bibitem{DBLP:conf/kdd/ZhangSHSC19}
C.~Zhang, D.~Song, C.~Huang, A.~Swami, and N.~V. Chawla, ``Heterogeneous graph
  neural network,'' in \emph{Proceedings of the 25th {ACM} {SIGKDD}
  International Conference on Knowledge Discovery {\&} Data Mining}.\hskip 1em
  plus 0.5em minus 0.4em\relax {ACM}, 2019, pp. 793--803.

\bibitem{DBLP:conf/www/0004ZMK20}
X.~Fu, J.~Zhang, Z.~Meng, and I.~King, ``{MAGNN:} metapath aggregated graph
  neural network for heterogeneous graph embedding,'' in \emph{{WWW} '20: The
  Web Conference 2020}.\hskip 1em plus 0.5em minus 0.4em\relax {ACM} / {IW3C2},
  2020, pp. 2331--2341.

\bibitem{DBLP:journals/pvldb/SunHYYW11}
Y.~Sun, J.~Han, X.~Yan, P.~S. Yu, and T.~Wu, ``Pathsim: Meta path-based top-k
  similarity search in heterogeneous information networks,'' \emph{{PVLDB}},
  vol.~4, no.~11, pp. 992--1003, 2011.

\bibitem{DBLP:conf/kdd/ShiCZG017}
Y.~Shi, P.~Chan, H.~Zhuang, H.~Gui, and J.~Han, ``Prep: Path-based relevance
  from a probabilistic perspective in heterogeneous information networks,'' in
  \emph{Proceedings of the 23rd {ACM} {SIGKDD}}.\hskip 1em plus 0.5em minus
  0.4em\relax {ACM}, 2017, pp. 425--434.

\bibitem{DBLP:conf/aaai/LiHW18}
Q.~Li, Z.~Han, and X.~Wu, ``Deeper insights into graph convolutional networks
  for semi-supervised learning,'' in \emph{Proceedings of the Thirty-Second
  {AAAI} Conference on Artificial Intelligence}.\hskip 1em plus 0.5em minus
  0.4em\relax {AAAI} Press, 2018, pp. 3538--3545.

\bibitem{DBLP:conf/aaai/ChenLLLZS20}
D.~Chen, Y.~Lin, W.~Li, P.~Li, J.~Zhou, and X.~Sun, ``Measuring and relieving
  the over-smoothing problem for graph neural networks from the topological
  view,'' in \emph{{AAAI} 2020}.\hskip 1em plus 0.5em minus 0.4em\relax {AAAI}
  Press, 2020, pp. 3438--3445.

\bibitem{bruna2013spectral}
J.~Bruna, W.~Zaremba, A.~Szlam, and Y.~LeCun, ``Spectral networks and locally
  connected networks on graphs,'' \emph{arXiv preprint arXiv:1312.6203}, 2013.

\bibitem{DBLP:conf/iclr/KipfW17}
T.~N. Kipf and M.~Welling, ``Semi-supervised classification with graph
  convolutional networks,'' in \emph{ICLR}.\hskip 1em plus 0.5em minus
  0.4em\relax OpenReview.net, 2017.

\bibitem{hamilton2017inductive}
W.~Hamilton, Z.~Ying, and J.~Leskovec, ``Inductive representation learning on
  large graphs,'' in \emph{Advances in neural information processing systems},
  2017, pp. 1024--1034.

\bibitem{dou2020enhancing}
Y.~Dou, Z.~Liu, L.~Sun, Y.~Deng, H.~Peng, and P.~S. Yu, ``Enhancing graph
  neural network-based fraud detectors against camouflaged fraudsters,'' in
  \emph{Proceedings of the 29th ACM International Conference on Information \&
  Knowledge Management}, 2020, pp. 315--324.

\bibitem{peng2019fine}
H.~Peng, J.~Li, Q.~Gong, Y.~Song, Y.~Ning, K.~Lai, and P.~S. Yu, ``Fine-grained
  event categorization with heterogeneous graph convolutional networks,''
  \emph{IJCAI}, 2019.

\bibitem{liu2020basket}
Z.~Liu, X.~Li, Z.~Fan, S.~Guo, K.~Achan, and P.~S. Yu, ``Basket recommendation
  with multi-intent translation graph neural network,'' \emph{arXiv preprint
  arXiv:2010.11419}, 2020.

\bibitem{li2020dynamic}
X.~Li, M.~Zhang, S.~Wu, Z.~Liu, L.~Wang, and P.~S. Yu, ``Dynamic graph
  collaborative filtering,'' in \emph{ICDM}, 2020.

\bibitem{gao2020hincti}
Y.~Gao, L.~Xiaoyong, P.~Hao, B.~Fang, and P.~Yu, ``Hincti: A cyber threat
  intelligence modeling and identification system based on heterogeneous
  information network,'' \emph{IEEE Transactions on Knowledge and Data
  Engineering}, 2020.

\bibitem{DBLP:journals/corr/abs-1710-10903}
P.~Velickovic, G.~Cucurull, A.~Casanova, A.~Romero, P.~Li{\`{o}}, and
  Y.~Bengio, ``Graph attention networks,'' in \emph{ICLR}, vol. abs/1710.10903,
  2018.

\bibitem{cao2020multi}
Y.~Cao, H.~Peng, and S.~Y. Philip, ``Multi-information source hin for medical
  concept embedding,'' in \emph{Pacific-Asia Conference on Knowledge Discovery
  and Data Mining}.\hskip 1em plus 0.5em minus 0.4em\relax Springer, 2020, pp.
  396--408.

\bibitem{DBLP:journals/tkde/ShiLZSY17}
\BIBentryALTinterwordspacing
C.~Shi, Y.~Li, J.~Zhang, Y.~Sun, and P.~S. Yu, ``A survey of heterogeneous
  information network analysis,'' \emph{{IEEE} Trans. Knowl. Data Eng.},
  vol.~29, no.~1, pp. 17--37, 2017. [Online]. Available:
  \url{https://doi.org/10.1109/TKDE.2016.2598561}
\BIBentrySTDinterwordspacing

\bibitem{DBLP:conf/iclr/ZhaoA20}
L.~Zhao and L.~Akoglu, ``Pairnorm: Tackling oversmoothing in gnns,'' in
  \emph{8th International Conference on Learning Representations, {ICLR} 2020},
  2020.

\bibitem{goldberger2000physiobank}
A.~L. Goldberger, L.~A. Amaral, L.~Glass, J.~M. Hausdorff, P.~C. Ivanov, R.~G.
  Mark, J.~E. Mietus, G.~B. Moody, C.-K. Peng, and H.~E. Stanley, ``Physiobank,
  physiotoolkit, and physionet: components of a new research resource for
  complex physiologic signals,'' \emph{Circulation}, vol. 101, no.~23, pp.
  e215--e220, 2000.

\bibitem{johnson2016mimic}
A.~E. Johnson, T.~J. Pollard, L.~Shen, H.~L. Li-wei, M.~Feng, M.~Ghassemi,
  B.~Moody, P.~Szolovits, L.~A. Celi, and R.~G. Mark, ``Mimic-iii, a freely
  accessible critical care database,'' \emph{Scientific data}, vol.~3, p.
  160035, 2016.

\bibitem{DBLP:conf/cikm/HosseiniCWSS18}
A.~Hosseini, T.~Chen, W.~Wu, Y.~Sun, and M.~Sarrafzadeh, ``Heteromed:
  Heterogeneous information network for medical diagnosis,'' in \emph{CIKM},
  2018, pp. 763--772.

\bibitem{DBLP:conf/vissym/RauberFT16}
P.~E. Rauber, A.~X. Falc{\~{a}}o, and A.~C. Telea, ``Visualizing time-dependent
  data using dynamic t-sne,'' in \emph{EuroVis}, 2016, pp. 73--77.

\bibitem{DBLP:conf/kdd/PerozziAS14}
B.~Perozzi, R.~Al{-}Rfou, and S.~Skiena, ``Deepwalk: online learning of social
  representations,'' in \emph{SIGKDD}, 2014, pp. 701--710.

\bibitem{DBLP:conf/kdd/DongCS17}
Y.~Dong, N.~V. Chawla, and A.~Swami, ``metapath2vec: Scalable representation
  learning for heterogeneous networks,'' in \emph{SIGKDD}, 2017, pp. 135--144.

\end{thebibliography}

\end{document}